\documentclass[runningheads]{llncs}
\usepackage{algorithm2e}
\usepackage{cite}
\usepackage{amsmath,amssymb,amsfonts}
\usepackage{algorithmic}
\usepackage{graphicx}
\usepackage{textcomp}
\usepackage{booktabs}
\usepackage{caption}  
\usepackage{subcaption}
\usepackage{multirow}
\usepackage{marvosym} 
\usepackage{color}
\definecolor{r}{rgb}{0,0,0}

\usepackage[T1]{fontenc}
\usepackage{graphicx,verbatim}
\title{Diverse Normal Prototypes-Guided Contrastive Reconstruction for Medical Anomaly Detection}
\titlerunning{DNP-ConFormer for Medical Anomaly Detection}
\begin{document}
% ================= Author (Anonymous for double-blind) =================
% \author{Anonymous Author(s)}
% \institute{Anonymous Institution(s)}
\author{Luhu Li \inst{1} \and
Bin Liu\inst{2} \and
Bowen Lin\inst{2} \textsuperscript{(\Letter)} \and
Zihan Shen\inst{1} \and
Chengwei Wang\inst{3} \and
Shujun Fu\inst{1} \textsuperscript{(\Letter)}}   %% Added for anonymized MICCAI submission

% index{Li, Luhu}
% index{Liu, Bin}
% index{Lin, Bowen}
% index{Shen, Zihan}
% index{Wang, Chengwei}
% index{Fu, Shujun}

\authorrunning{Li et al.}
\institute{School of Mathematics, Shandong University, Jinan 250100, China \\
    \email{shujunfu@163.com} \and Department of Interventional Medicine and Minimally Invasive Oncology, The Second Qilu Hospital of Shandong University, Jinan 250033, China \\
    \email{linbowencore7@outlook.com} \and Department of Neurosurgery, The Second Qilu Hospital of Shandong University, Jinan 250033, China}

\maketitle

% \begin{abstract}
% 	Anomaly detection in medical images is challenging due to limited annotations and a domain gap compared to natural images. Existing reconstruction methods often rely on frozen pre-trained encoders, which limits adaptation to domain-specific features and reduces localization accuracy.
% 	Prototype-based learning offers interpretability and clustering benefits but suffers from {prototype collapse}, where few prototypes dominate training, harming diversity and generalization.
% 	To address these challenges, we propose a unified framework combining a \textit{trainable encoder with prototype-guided reconstruction} and a novel \textit{Diversity-Aware Alignment Loss}. The trainable encoder, enhanced by a \textit{momentum branch}, enables stable domain-adaptive feature learning. A lightweight {Prototype Extractor} mines informative normal prototypes to guide the decoder via attention for precise reconstruction. Our diversity-aware alignment objective guides training towards balanced feature-to-prototype assignments, effectively preventing prototype collapse.
% 	Experiments on multiple medical imaging benchmarks show significant improvements in representation quality and anomaly localization, outperforming prior methods. Visualizations and prototype assignment analyses further validate the effectiveness of our anti-collapse mechanism and enhanced interpretability. The code will be released upon acceptance.
	
% \end{abstract}

\begin{abstract}
Anomaly detection in medical images is challenging due to limited annotations and the domain gap. Existing reconstruction-based methods often rely on frozen pre-trained encoders, restricting adaptation to domain-specific patterns and degrading localization accuracy. Meanwhile, prototype-based learning offers interpretable representations but commonly suffers from {prototype collapse}, where a few prototypes dominate training and reduce diversity.
To address these issues, we propose {DNP-ConFormer}, a unified framework that integrates a trainable encoder with prototype-guided reconstruction and a {Diversity-Aware Alignment Loss}. A momentum encoder enables stable domain-adaptive representation learning, while a lightweight {Prototype Extractor} discovers informative normal prototypes and injects them into the decoder via attention to guide reconstruction. The proposed alignment objective further encourages balanced feature-to-prototype assignments, effectively mitigating prototype collapse.
Extensive experiments on multiple medical imaging benchmarks demonstrate improved representation quality and anomaly localization compared with prior methods. Visualization and prototype assignment analyses further validate the effectiveness and interpretability of our approach. The code is available at \url{https://github.com/liluhu0/DNP-ConFormer}.
\end{abstract}


\section{Introduction}

\label{introduction}
Unsupervised anomaly detection (UAD) provides an alternative to supervised deep learning for medical image analysis, where obtaining pixel-level annotations is costly and anomalies are often rare~\cite{Unet, Dosovitskiy2020, 10891867, Baid2021, Nguyen2022}. By training only on normal images, UAD methods detect anomalies as deviations at inference time, which can improve scalability for tasks such as early screening~\cite{RD4AD}.

\begin{figure}[!t]
	\centering
	\begin{subfigure}[t]{0.48\columnwidth}
		\centering
		\includegraphics[width=0.9\linewidth]{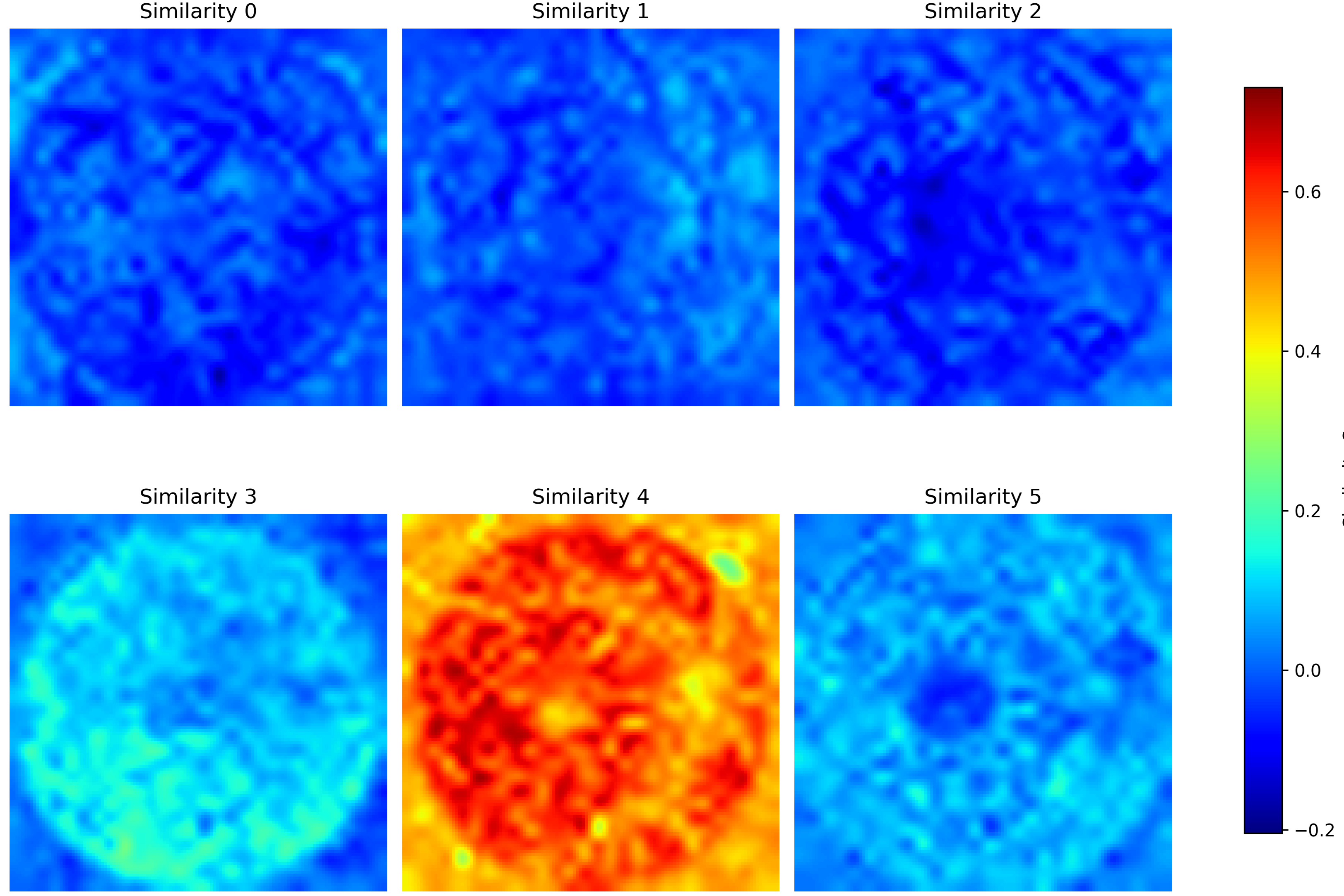}
		\caption{INP-Former}
		\label{Compare_INP_loss-a}
	\end{subfigure}
	\hfill
	\begin{subfigure}[t]{0.48\columnwidth}
		\centering
		\includegraphics[width=0.9\linewidth]{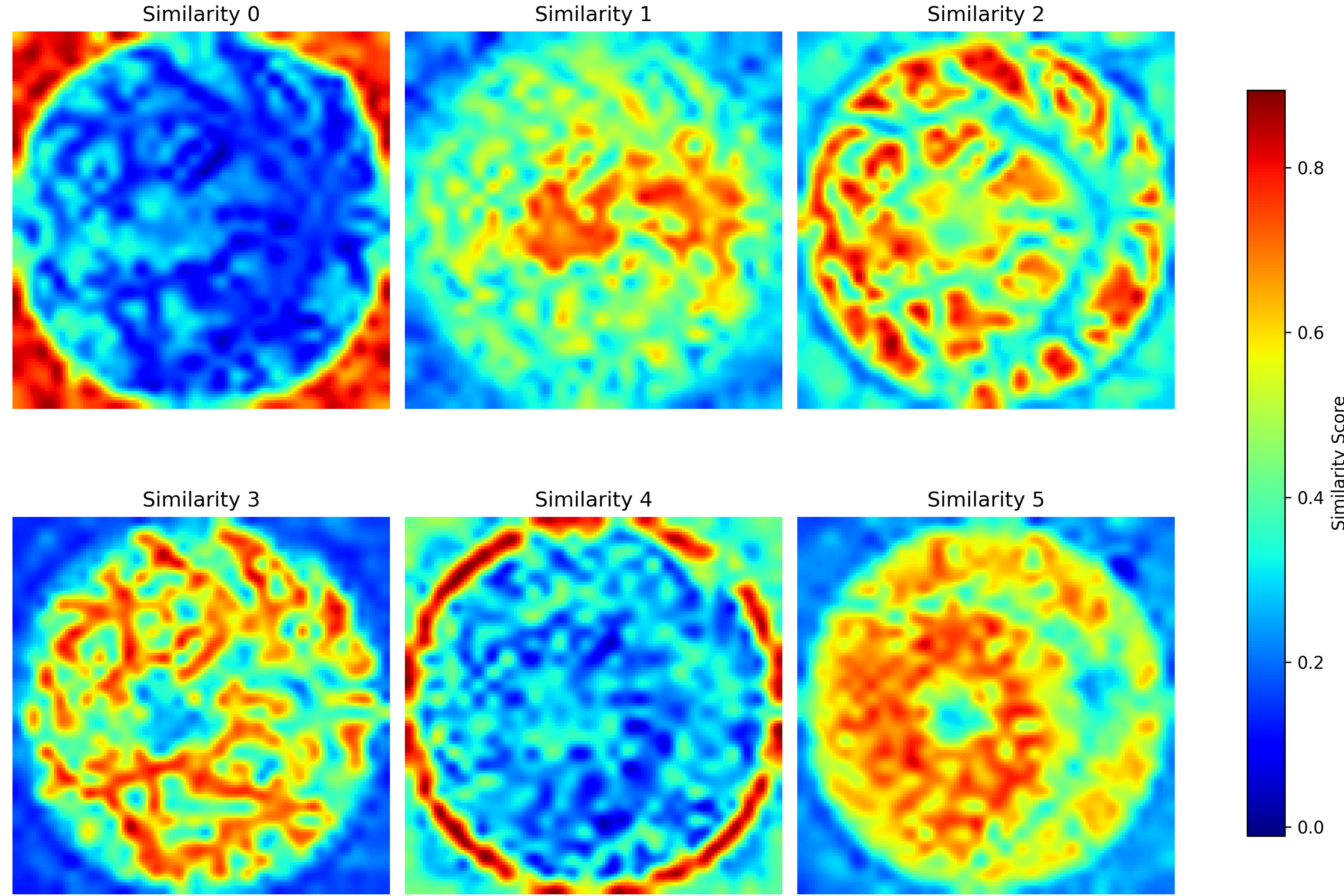}
		\caption{Ours}
		\label{Compare_INP_loss-b}
	\end{subfigure}
%	\caption{Visualizations of prototypes on a fundus image:
%		(a) Results from INP-Former using the original coherence loss, where all patch features are most similar to the fifth prototype, indicating a collapse into a single dominant prototype.
%		(b) Results from our method using the Diversity-Aware Alignment loss, where the prototypes effectively capture diverse semantic information across different patches.}
	\caption{Prototype visualizations on fundus image: (a) INP-Former with coherence loss suffers prototype collapse; (b) Our method with Diversity-Aware Alignment loss achieves diverse semantic assignments.}
	\label{Compare_INP_loss}
\end{figure}

Among these approaches, reconstruction-based models assume anomalies yield higher reconstruction errors. However, image-level strategies often fail to capture fine-grained local anomalies. Recent teacher-student reverse distillation frameworks address this by training a student decoder to reconstruct multi-scale features from a frozen teacher encoder~\cite{RD4AD}. Representations produced by the teacher are further refined using contrastive learning (CL)~\cite{ReContrast, EDC, EA2D}. Notably, self-supervised Vision Transformers (ViTs) provide richer semantic features for anomaly localization than CNNs or supervised ViTs~\cite{DINO, DINOv2, DINOv2_reg}.
Building on these advances, INP-Former~\cite{INP-Former} employs a self-supervised ViT teacher to extract Intrinsic Normal Prototypes (INPs) that guide student reconstruction via a coherence objective. While effective in industrial inspection, two challenges remain when applying such methods to medical images: (1) limited domain adaptability of the frozen encoder when the source (natural images) and target (medical images) domains differ substantially, and (2) prototype collapse under the coherence objective, where low-contrast medical features tend to converge to a dominant prototype, reducing the diversity of normal representations (Fig.~\ref{Compare_INP_loss-a},~\ref{Compare_INP_loss-b}).

In this work, we propose \textbf{DNP-ConFormer} (Diverse Normal Prototypes-guided Contrastive Reconstruction), a dual-branch framework designed for domain-adaptive anomaly detection. Unlike reverse distillation approaches that rely on a frozen teacher encoder, our framework employs a momentum-updated teacher that evolves during training. The student branch combines a trainable encoder with a DNP-guided decoder, where Diverse Normal Prototypes (DNPs) extracted from encoder features are injected via cross-attention to guide reconstruction. The teacher branch provides relatively stable reference features through exponential moving average updates. Anomaly maps are derived from feature discrepancies between the two branches, which helps improve robustness under domain shifts. 
To mitigate prototype collapse, we consider prototype learning from a long-tail perspective by treating each prototype as a class center and patch-level features as samples. Based on this observation, we introduce a Diversity-Aware Alignment (DAA) loss that encourages balanced feature–prototype assignments while maintaining inter-prototype distinctiveness.

\begin{figure*}[!t]
	\includegraphics[width=\textwidth]{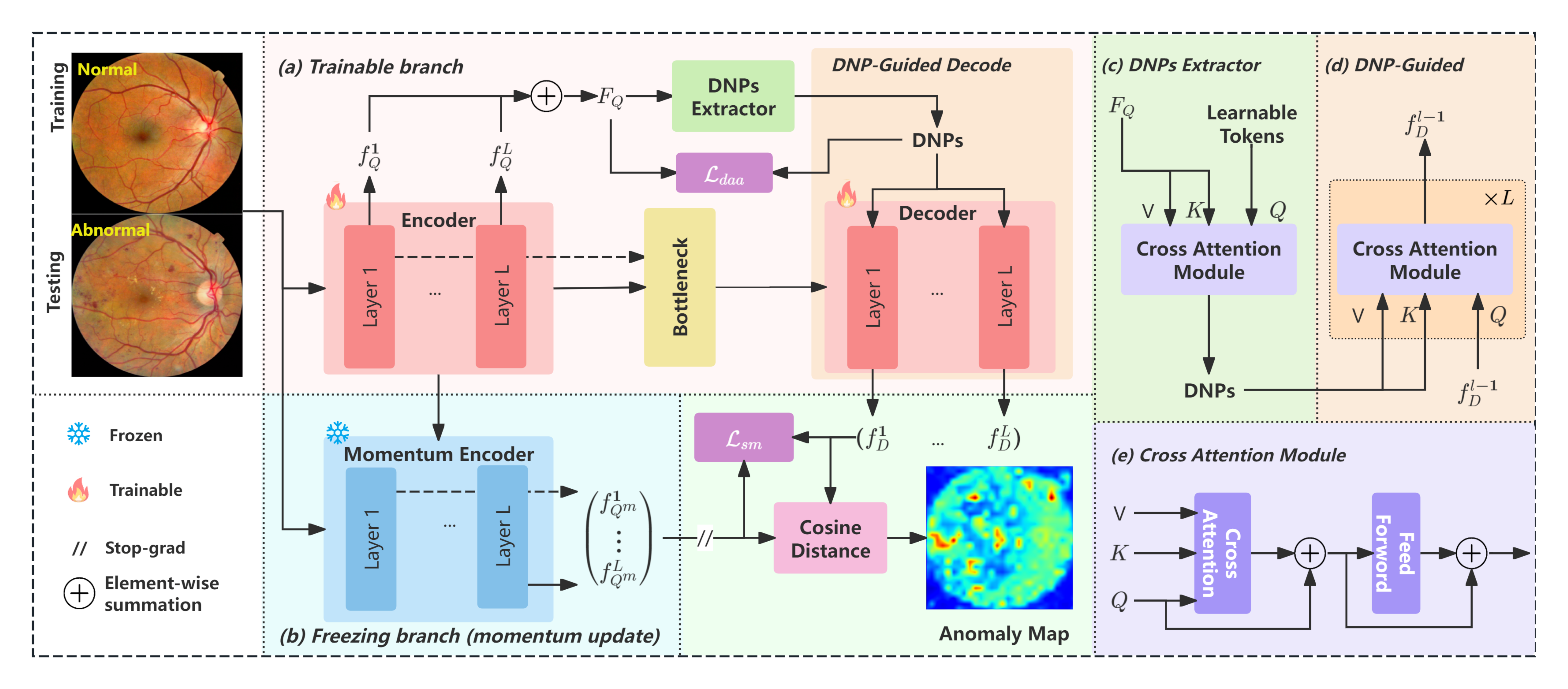}
	\caption{
		Overview of the proposed {DNP-ConFormer}.
		(a) The {trainable student branch}. 
		(b) The {momentum teacher branch}. 
		(c) The {DNP extraction module}. 
		(d) The {DNP-guided decoding module}. 
		(e) The {cross-attention module}.
	}
	
	\label{pipeline}
\end{figure*}

\section{Method}
\label{sec:method}

We present the overall pipeline of \textbf{DNP-ConFormer}, illustrated in Fig.~\ref{pipeline}. The framework consists of a \textbf{student branch} for representation learning and reconstruction, and a \textbf{teacher branch} that provides stable reference features via momentum updating. Given an input image $\mathbf{I}\in\mathbb{R}^{H\times W\times 3}$, the student encoder $\mathcal{Q}$ extracts multi-scale features $f_{\mathcal{Q}}=\{f_{\mathcal{Q}}^1,\dots,f_{\mathcal{Q}}^L\}$ with $f_{\mathcal{Q}}^l\in\mathbb{R}^{N\times C}$, which are aggregated into a unified representation $\mathbf{F}_{\mathcal{Q}}=\frac{1}{L}\sum_{l=1}^{L}f_{\mathcal{Q}}^{l}$. To capture diverse normal patterns, we introduce a set of $M$ learnable tokens $\{t_1,\dots,t_M\}$ and employ a cross-attention module to extract \textbf{Diverse Normal Prototypes} $\mathbf{P}=\mathcal{E}(\mathbf{F}_{\mathcal{Q}},\{t_m\})$, where $\mathbf{P}\in\mathbb{R}^{M\times C}$ represents different modes of normality in the feature space. These prototypes are injected into the DNP-guided decoder $\mathcal{D}$ via cross-attention with the bottleneck representation $F_{\mathcal{B}}=\mathcal{B}(f_{\mathcal{Q}})$, where $\mathcal{B}$ denotes a bottleneck aggregation module, enabling the decoder to reconstruct feature representations aligned with normal patterns. The \textbf{teacher branch} maintains a momentum encoder $\mathcal{Q}^{m}$ whose parameters $\theta_{\mathcal{Q}^{m}}$ are updated as an exponential moving average (EMA) of the student encoder parameters $\theta_{\mathcal{Q}}$, producing temporally smoothed and stable reference features. During training, we compute the cosine distance between student and teacher features to measure representation discrepancy, which is used to generate the anomaly map at the patch level. The overall training objective consists of two components: a \textbf{soft-mining loss} $\mathcal{L}_{sm}$ that enforces consistency between the student and teacher representations while emphasizing informative samples, and a \textbf{Diversity-Aware Alignment loss} $\mathcal{L}_{daa}$ that encourages balanced and discriminative prototype assignments.

\subsection{Encoder Adaptation Frameworks}
\label{sec:encoder-adapt}
\begin{figure}[!t]
	\centering
	\begin{subfigure}[t]{0.5\columnwidth}
		\centering
		\includegraphics[width=\textwidth]{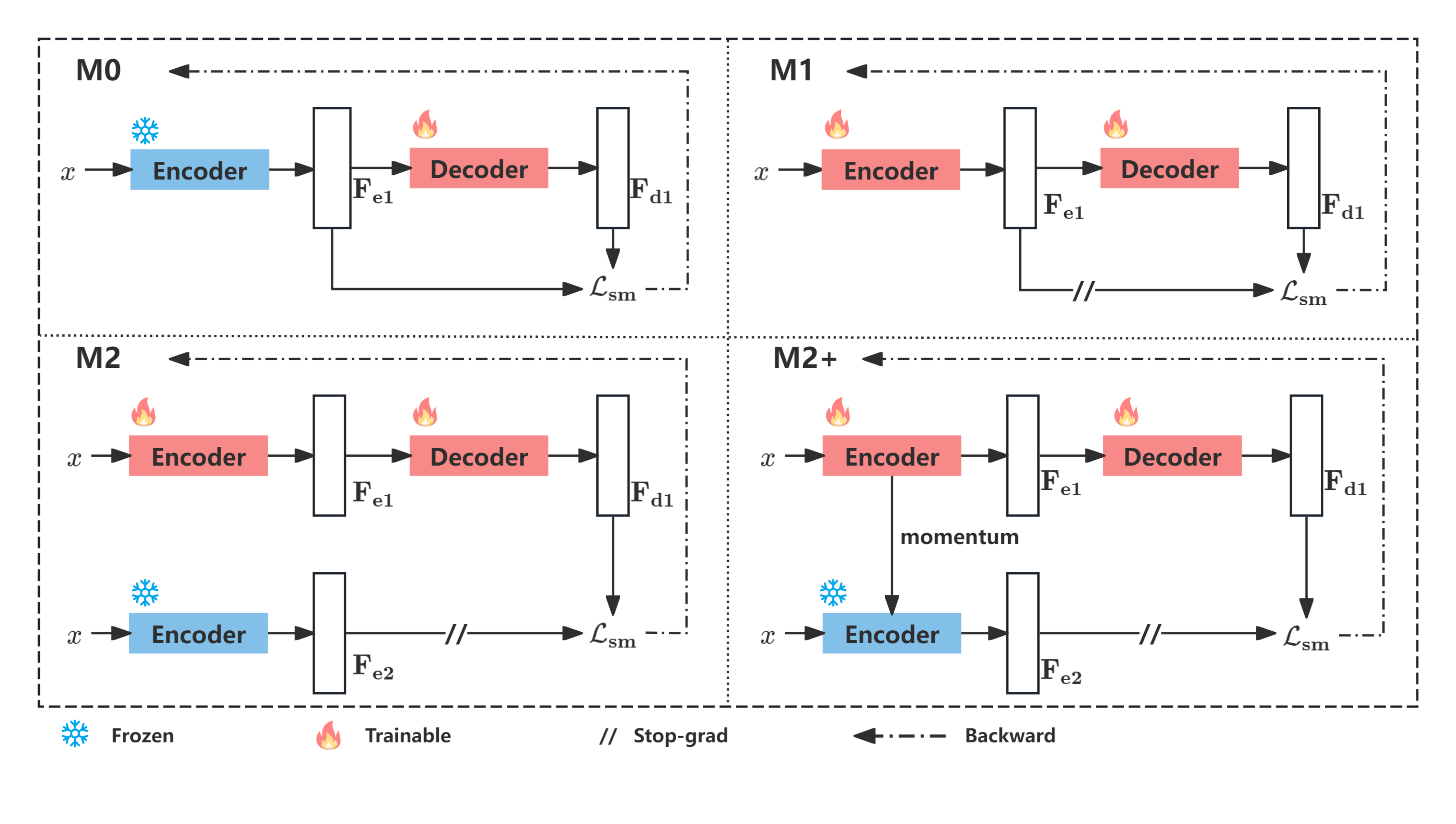}
		\label{train_encoder}
	\end{subfigure}
	\hfill
	\begin{subfigure}[t]{0.45\columnwidth}
		\centering
		\includegraphics[width=\textwidth]{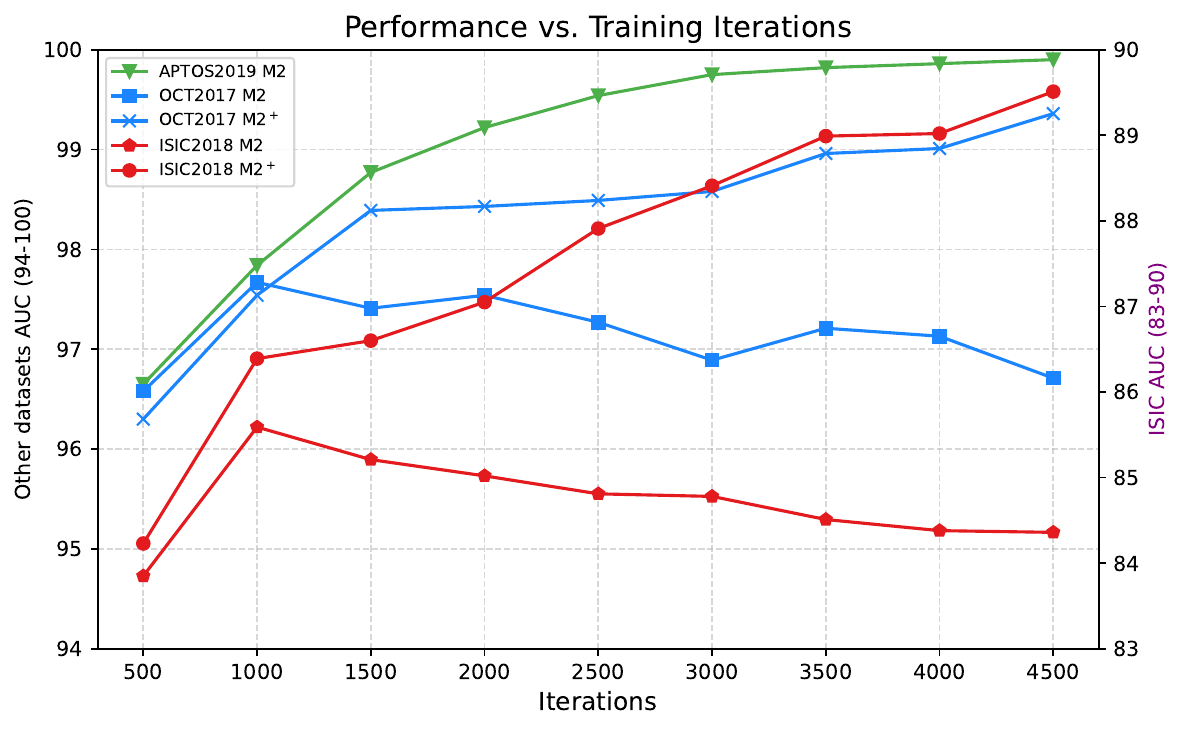}
		\label{Performance_vs_Training_Iterations}
	\end{subfigure}
	\caption{(a) Model variants (M0 to M$2^+$): Progressive design for domain adaptation in anomaly detection. (b) Performance trends over training iterations on different datasets. Under the M2 framework, performance steadily improves on APTOS2019, while OCT2017 peaks at 1000 iterations before declining. With the M2$^+$ framework, OCT2017 also shows continuous improvement.}
	\label{fig:encoder_adaptation}
\end{figure}

Frozen natural-image encoders $\mathcal{Q}$ lack adaptability to medical imaging data, leading to domain inflexibility, representation misalignment, and uncorrectable reconstruction errors. A direct approach is to unfreeze $\mathcal{Q}$ and jointly optimize it with the decoder $\mathcal{D}$ (\textbf{M1} in Fig.~\ref{fig:encoder_adaptation}). \textcolor{r}{However, with limited medical data, end-to-end fine-tuning often causes gradient instability, overfitting, and catastrophic forgetting of useful pretrained priors.} Inspired by SimSiam \cite{Simsiam}, we adopt an asymmetric twin-path framework (\textbf{M2} in Fig.~\ref{fig:encoder_adaptation}) for stable adaptation, comprising a trainable online path ($x \rightarrow \mathcal{Q} \rightarrow \mathcal{B} \rightarrow \mathcal{D} \Rightarrow \hat{f}_{\mathcal{D}}$) and a stable frozen target path ($x \rightarrow \mathcal{Q}^{\text{frz}} \Rightarrow f_{\mathcal{Q}^{\text{frz}}}$). 
This asymmetric design mitigates representation collapse and enables gradual alignment between the online and target representations via the following objective:
\begin{equation}
	\small
	\mathcal{L}_{sm}^{\text{M2}} = 1 - \cos\left( \operatorname{sg}(f_{\mathcal{Q}^{\text{frz}}}), \hat{f}_{\mathcal{D}} \right).
	\label{L_sm_M2}
\end{equation}

Nevertheless, the statically frozen target encoder cannot evolve with the online network, which may limit adaptation capacity and lead to performance saturation. To overcome this limitation, we further propose \textbf{M$2^+$} (Fig.~\ref{fig:encoder_adaptation}), where the target encoder is replaced with a momentum encoder $\mathcal{Q}^{m}$ updated using an exponential moving average:
\begin{equation}
	\small
	\theta_{\mathcal{Q}^{m}} \leftarrow \beta \theta_{\mathcal{Q}^{m}} + (1 - \beta) \theta_{\mathcal{Q}}.
\end{equation}
This design provides a slowly evolving target representation that improves temporal consistency, preserves pretrained knowledge, and stabilizes optimization, thereby enabling stable convergence across diverse medical datasets as illustrated in Fig.~\ref{fig:encoder_adaptation}. Accordingly, we adopt M$2^+$ as our final design, and the training objective is defined as
\begin{equation}
	\small
	\mathcal{L}_{sm}^{\text{M$2^+$}} = 1 - \cos\left( \operatorname{sg}(f_{\mathcal{Q}^{m}}), \hat{f}_{\mathcal{D}} \right).
\end{equation}

\subsection{Diversity-Aware Normal Prototypes}

\textbf{Prototype collapse.} 
\textcolor{r}{The baseline coherence objective often leads to single-prototype dominance, where most patches are assigned to one prototype while the others remain underutilized, thereby limiting the representational capacity of the model.} This issue is particularly pronounced in medical images due to their low contrast and limited texture variability. The collapse arises because the consistency objective enforces feature alignment but does not impose any constraint on prototype diversity; consequently, early advantages gained by certain prototypes become self-reinforcing, resulting in imbalanced prototype utilization.

\textbf{Diversity-Aware Alignment Loss: A Long-Tail Learning Perspective.}
From a long-tail learning perspective, prototype collapse resembles class-imbalance behavior, where a small subset of prototypes dominates feature assignments. Existing consistency objectives focus on patch-level alignment but overlook prototype-level distribution imbalance. To address this issue, we encourage balanced participation across prototypes by promoting diverse and informative feature alignment, leading to the proposed \textbf{Diversity-Aware Alignment Loss}. For each prototype $p_j$, we define the assigned feature set $\mathcal{M}_j$ as
\begin{equation}
	\small
	\mathcal{M}_{j} = \left\{ i \in \{1, \ldots, N\} \ \middle| \ 
	\arg\min_{m} \mathcal{S}(\mathbf{F}_{\mathcal{Q}}(i), p_m) = j 
	\right\}.
	\label{Mj_def}
\end{equation}

Each patch feature is assigned to its nearest prototype, and a distance-based loss is computed within each prototype group. To ensure robustness, we only include prototypes that have been assigned at least one feature. The Diversity-Aware Alignment Loss is defined as
\begin{equation} 
	\mathcal{L}_{\text{daa}} = \frac{1}{M} \sum_{j=1}^{M} 
	\mathbf{1}_{|\mathcal{M}_j| > 0} \cdot 
	\frac{1}{|\mathcal{M}_{j}|} \sum_{i \in \mathcal{M}_{j}} 
	\mathcal{S}(\mathbf{F}_\mathcal{Q}(i), p_j).
	\label{DAAloss}
\end{equation}

\textcolor{r}{This formulation promotes a more balanced assignment distribution across prototypes, analogous to maximizing entropy in prototype utilization, thereby mitigating collapse.} The indicator function $\mathbf{1}_{|\mathcal{M}_j| > 0}$ ensures that only non-empty prototype groups contribute to the loss, avoiding invalid gradients. The final objective is defined as
\begin{equation}
	\mathcal{L}_{\text{total}} = \mathcal{L}_{sm}^{\text{M}2^+} + \lambda \mathcal{L}_{daa}.
\end{equation}

\section{Experiments and Results}
\label{Experiments}
\subsection{\textcolor{r}{Experimental Setup}}
\label{Experimental setup}

%\textbf{Dataset.}We evaluate our method on three public medical imaging benchmarks: \textbf{OCT2017}~\cite{OCT2017}, a retinal OCT dataset with four categories where only the normal class is treated as healthy; \textbf{APTOS2019}~\cite{APTOS2019}, a color fundus dataset annotated with diabetic retinopathy grades 0–4, using grade 0 as normal; and \textbf{ISIC2018}~\cite{ISIC2018}, a skin lesion dataset with seven classes, where the nevus (NV) category is regarded as normal. For all public datasets, images are resized to $256\times256$ and center-cropped, with a crop size of $252\times252$ for OCT2017 and APTOS2019, and $224\times224$ for ISIC2018.To assess clinical applicability, we further evaluate on an in-house brain MIR dataset (\textbf{BrainMIR}) from routine clinical examinations, containing normal and tumor slices from multiple patients. Experiments follow the standard anomaly detection setting, with training on normal samples only and testing on mixed normal and abnormal cases. All images are resized to $256\times256$ and center-cropped to $224\times224$.

\textbf{Datasets.} We evaluate our method on three public benchmarks and one in-house clinical collection: \textbf{OCT2017}~\cite{OCT2017} ,  \textbf{APTOS2019}~\cite{APTOS2019},  \textbf{ISIC2018}~\cite{ISIC2018}, and an in-house brain MIR dataset (\textbf{BrainMIR}). 
%Following the standard anomaly detection protocol, models are trained exclusively on normal samples and tested on mixed normal/abnormal cases. 
All images are resized to $224\times224$.

\textbf{\textcolor{r}{Metrics}.} 
{We evaluate performance using image-level and pixel-level metrics. Image-level performance is measured by AUC, F1-score, and accuracy (ACC), with AUC as the primary, threshold-independent metric and F1/ACC computed at the F1-optimal threshold, while pixel-level evaluation on the BrainMIR dataset reports AUC, Average Precision (AP), pixel-level F1, and Area Under the Per-Region Overlap curve (AUPRO) for anomaly localization.}

{\textbf{\textcolor{r}{Implementation details.}}}
%ViT-Small/14 with DINOv2-R \cite{DINOv2_reg} weights is used as the default teacher model and the student decoder consists of eight Transformer layers. All experiments are conducted using a single NVIDIA RTX 3090 GPU under the PyTorch framework. The batch size is set to 32. We adopt the StableAdamW optimizer \cite{StableAdamW}, with separate learning rates assigned to different modules: 0.0001 for the decoder, bottleneck, and DNPs extractor, and 0.00001 for the encoder. The number of iterations is set to 8,000 for OCT2017, 5,000 for APTOS and BrainMIR, and 4,000 for ISIC2018. By default, the number of DNPs, the momentum $\beta$, and the weight of DAA loss $\lambda$ are set to 6, 0.9999, and 0.2, respectively. All reported results are averaged over five independent runs. More details are in the appendix.
{We employ ViT-Small/14 with DINOv2-R \cite{DINOv2_reg} weights as the teacher and an 8-layer Transformer as the student decoder. Models are trained on a single NVIDIA RTX 3090 using StableAdamW \cite{StableAdamW} with a batch size of 32. Learning rates are set to $10^{-4}$ for the decoder, bottleneck, and DNPs extractor, and $10^{-5}$ for the encoder. Training iterations vary by dataset: 8k (OCT2017), 5k (APTOS, BrainMIR), and 4k (ISIC2018). Hyperparameters are fixed at $N_{DNP}=6$, $\beta=0.999$, and $\lambda=0.2$. Results are averaged over five runs; see Appendix for details.}

% \subsection{Main Result}

% We evaluate the performance of \textbf{DNP-ConFormer} against a broad range of state-of-the-art methods across four benchmark medical anomaly detection datasets: APTOS2019, OCT2017, ISIC2018 and BrainMIR. The results of these methods are obtained by quoting directly from published papers or by reproducing the code. Experimental results are averaged over five runs, and Table \ref{tab:Performance comparison} shows the comparative results, where  \textbf{Bold} indicates the best result and \underline{underlined} indicates the second-best.

\subsection{\textcolor{r}{Main Result}}
We evaluate DNP-ConFormer against state-of-the-art (SOTA) methods on three public benchmarks (APTOS2019, OCT2017, ISIC2018) and an in-house BrainMIR dataset. All experimental results are averaged over five independent runs. Table~\ref{tab:Performance_comparison} and Table~\ref{tab:Performance_comparison_clinical} report the comparative results, where \textbf{bold} and \underline{underlined} indicate the best and second-best performance, respectively. Quantitative results and qualitative visualizations demonstrate the superiority and robustness of our model across diverse modalities.

\textbf{Performance Comparison.}
As shown in Table~\ref{tab:Performance_comparison}, DNP-ConFormer demonstrates superior performance, achieving the top-ranked AUC, F1-score, and ACC across all public benchmarks, outperforming the runner-up Recontrast on APTOS2019 and reaching near-saturated results on OCT2017. For the challenging ISIC2018, our method significantly exceeds competitors, confirming its strong generalization capability.
On the clinical BrainMIR dataset (Table~\ref{tab:Performance_comparison_clinical}), DNP-ConFormer demonstrates exceptional robustness. It leads in image-level detection (92.52\% AUC, 87.10\% F1-score, and 85.39\% ACC) over Dinomaly and significantly outperforms INP-Former in pixel-level localization across all metrics, including  98.53\% AUC, 46.03\% AP, 51.10\% F1-score, and 89.27\% AUPRO. These results highlight the model's effectiveness in real-world clinical scenarios.

%Table~\ref{tab:Performance_comparison_clinical} further evaluates the performance on the in-house \textbf{BrainMIR} dataset. DNP-ConFormer demonstrates strong robustness in this clinical setting, achieving clear improvements over state-of-the-art competitors \cite{Performance_comparison_clinical}. 
% Specifically, for image-level anomaly detection, our method achieves 92.52\% AUC, 87.10\% F1-score, and 85.39\% ACC, surpassing the second-best Dinomaly \cite{Performance_comparison_clinical}. 
% In terms of pixel-level anomaly localization, DNP-ConFormer significantly outperforms INP-Former across all metrics, reaching an AUC of 98.53\%, AP of 46.03\%, F1-score of 51.10\%, and AUPRO of 89.27\% \cite{Performance_comparison_clinical}.

\begin{figure}[!t]
	\centering
	\begin{subfigure}[t]{0.32\columnwidth}
		\centering
		\includegraphics[width=\linewidth, height=0.3\textheight, keepaspectratio]{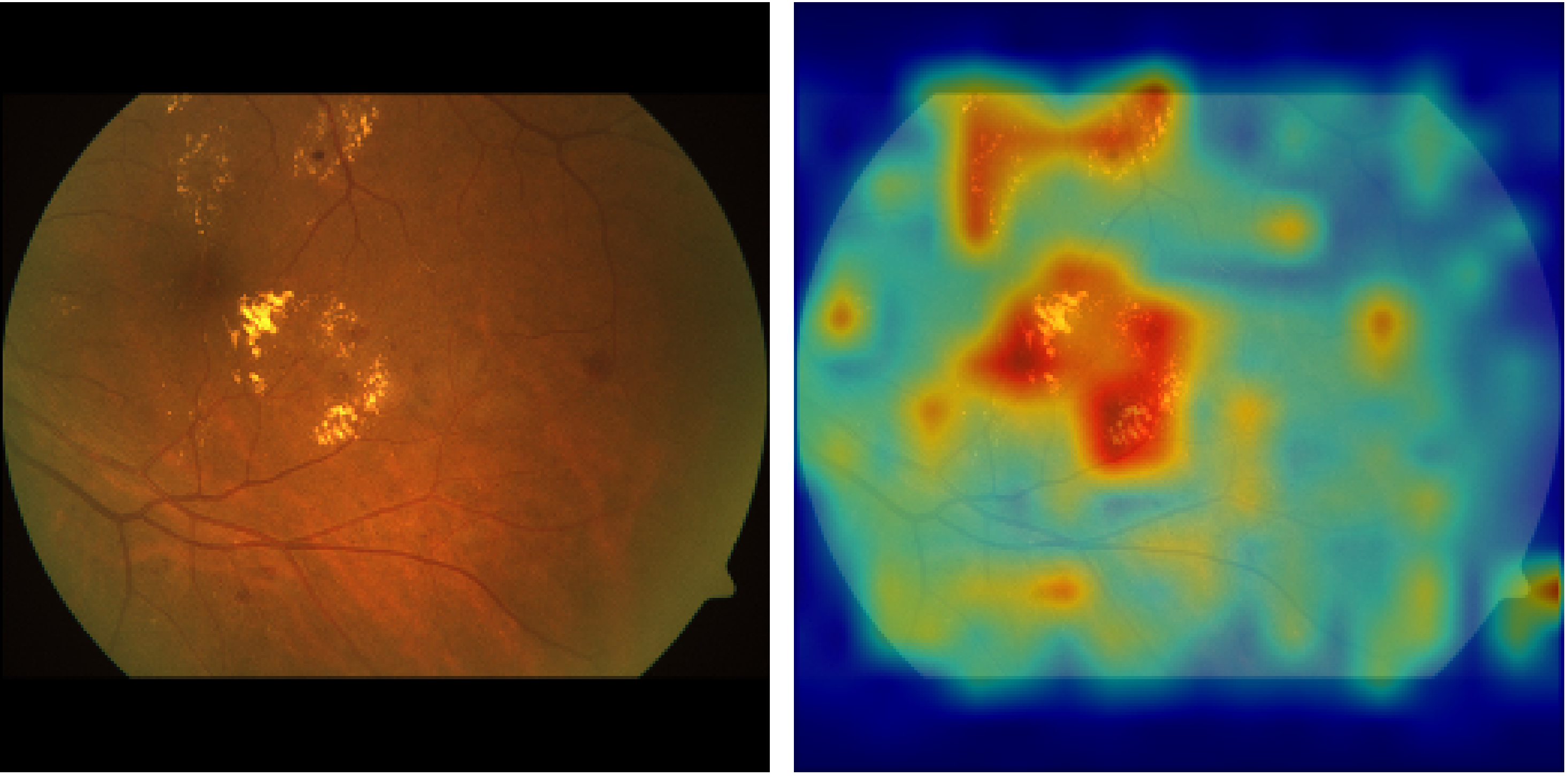}
		\caption{APTOS2019}
		\label{visualization_APTOS2019}
	\end{subfigure}
	\hfill
	\begin{subfigure}[t]{0.32\columnwidth}
		\centering
		\includegraphics[width=\linewidth, height=0.3\textheight, keepaspectratio]{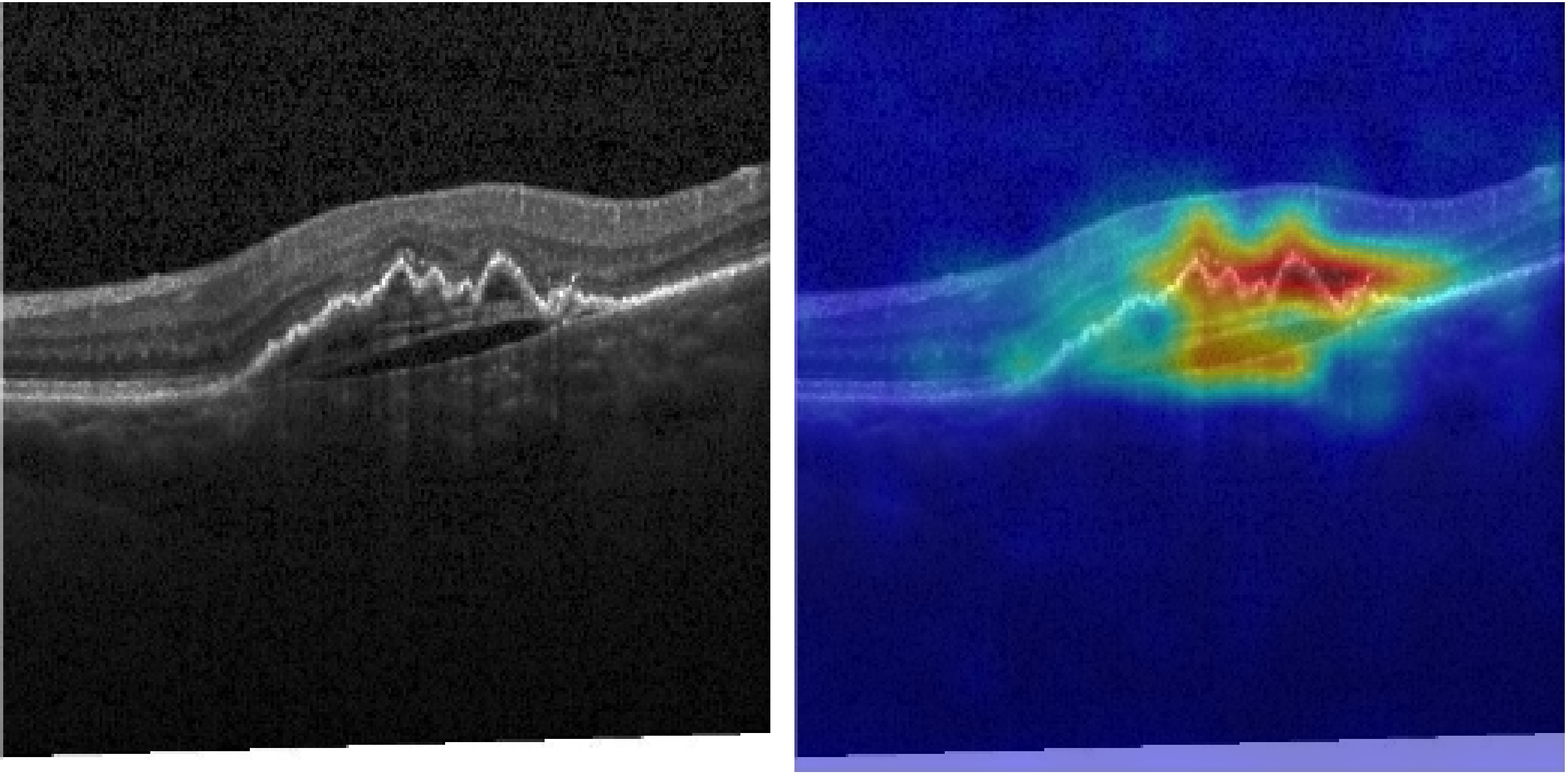}
		\caption{OCT2017}
		\label{visualization_OCT2017}
	\end{subfigure}
	\hfill
	\begin{subfigure}[t]{0.32\columnwidth}
		\centering
		\includegraphics[width=\linewidth, height=0.3\textheight, keepaspectratio]{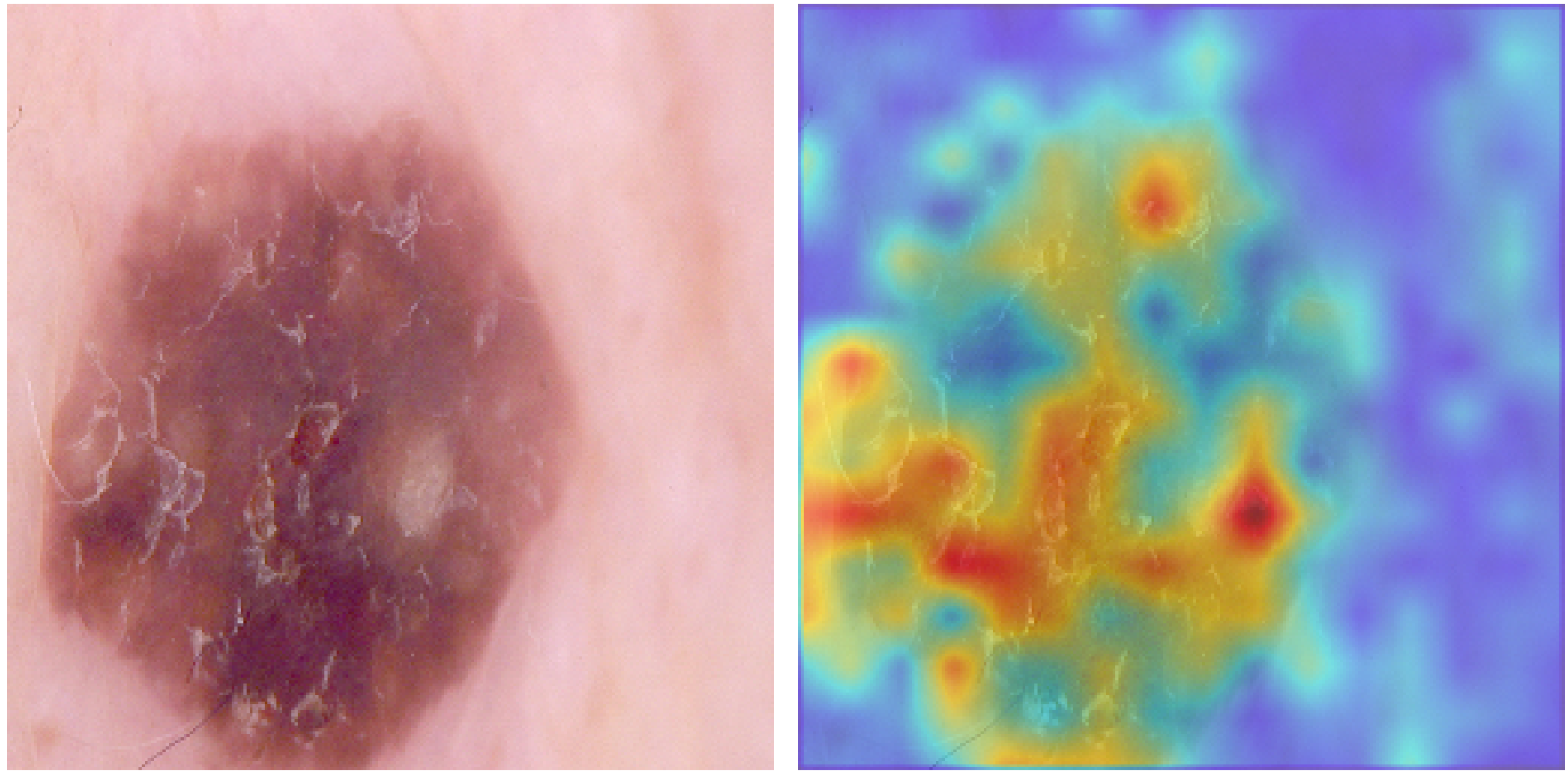}
		\caption{ISIC2018}
		\label{visualization_ISIC2018}
	\end{subfigure}
	\caption{Visualization of anomaly maps on three public benchmarks.}
	\label{fig:visualization}
\end{figure}

\begin{figure*}[!t]
	\centering
	\begin{subfigure}[t]{0.32\linewidth}
		\centering
		\includegraphics[width=\linewidth]{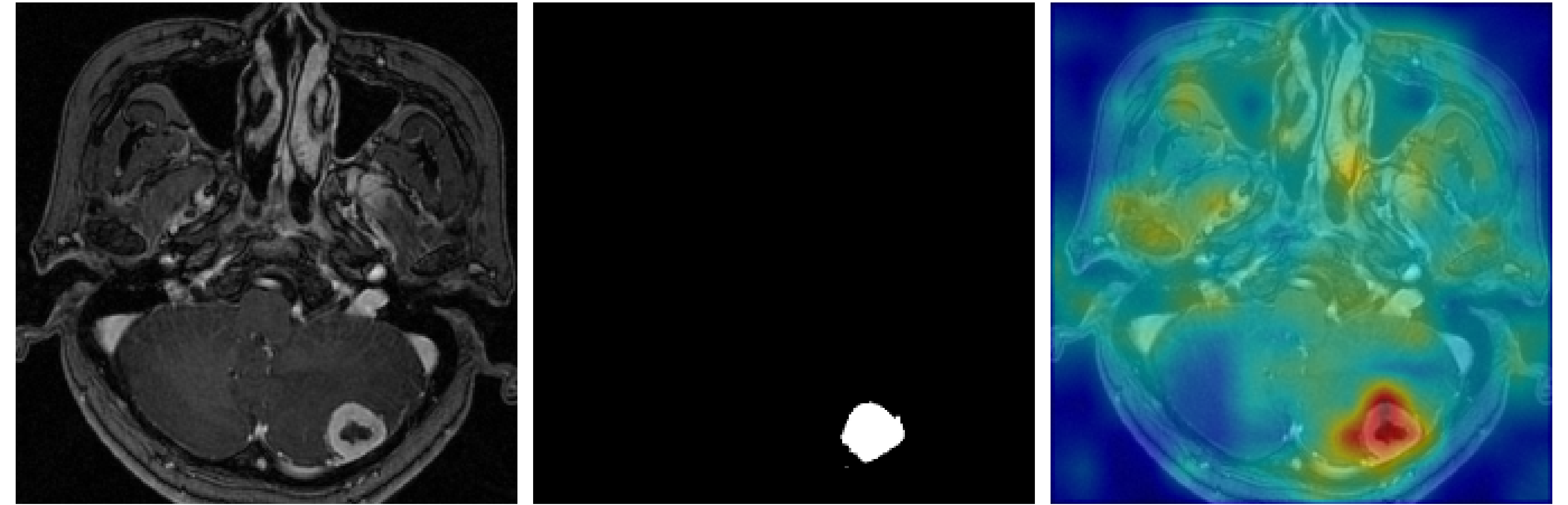}
		\caption{}
	\end{subfigure}
	\hfill
	\begin{subfigure}[t]{0.32\linewidth}
		\centering
		\includegraphics[width=\linewidth]{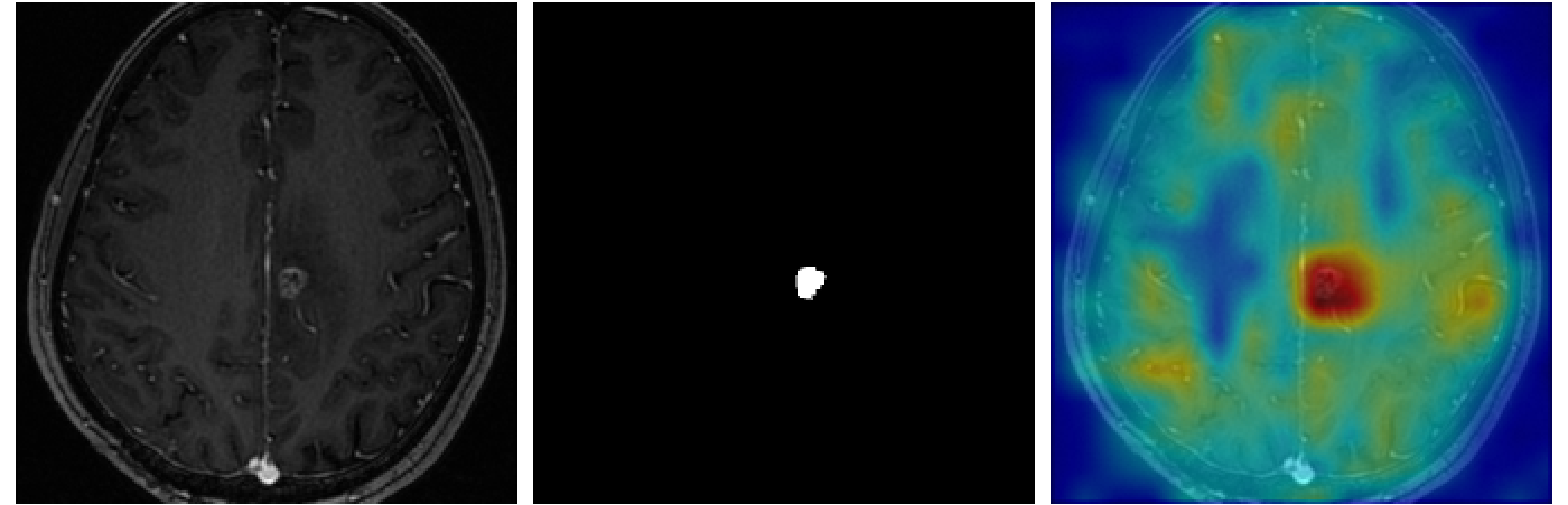}
		\caption{}
	\end{subfigure}
	\hfill
	\begin{subfigure}[t]{0.32\linewidth}
		\centering
		\includegraphics[width=\linewidth]{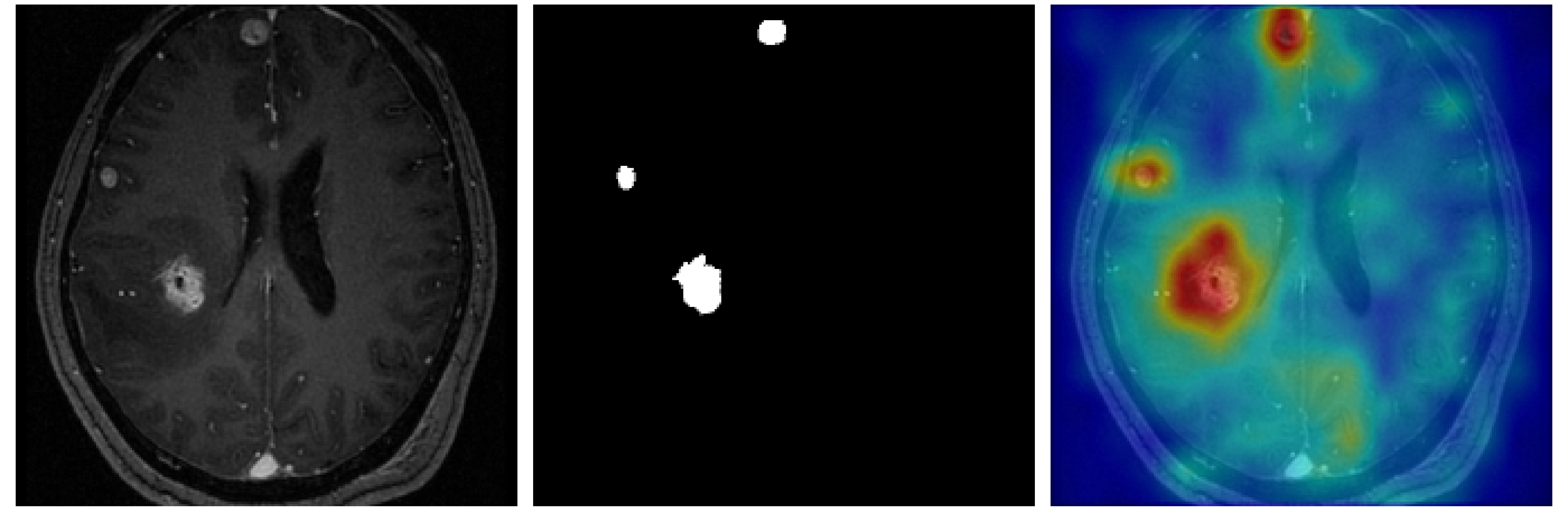}
		\caption{}
	\end{subfigure}
	\caption{Visualization of anomaly maps on the in-house BrainMIR dataset.}
	\label{fig:visualization_clinical}
\end{figure*}

\textbf{Qualitative Analysis.}
To assess localization precision, Fig.~\ref{fig:visualization} and Fig.~\ref{fig:visualization_clinical} visualize anomaly maps across various modalities. Despite structural complexities, DNP-ConFormer consistently generates focused heatmaps that closely align with pathological regions while suppressing background noise. Particularly in BrainMIR, the high-response regions show a high degree of overlap with pixel-level ground truth, confirming the model's ability to learn clinically meaningful representations.

\begin{table}[!h]
	\centering
	\caption{\textcolor{r}{Performance comparison on public benchmarks.}}
	\label{tab:Performance_comparison}
	\footnotesize
	\setlength{\tabcolsep}{1mm}
	%\resizebox{\columnwidth}{!}{%等比例缩放
		\begin{tabular}{@{}l |ccc|ccc|ccc@{}}
			\toprule
			{Dataset $\rightarrow$} & \multicolumn{3}{|c|}{APTOS2019} & \multicolumn{3}{c|}{OCT2017} &\multicolumn{3}{c}{ISIC2018} \\
			\cmidrule(lr){2-4} \cmidrule(lr){5-7} \cmidrule(lr){8-10}
			{Method $\downarrow$}& AUC & F1 & ACC & AUC & F1 & ACC & AUC & F1 & ACC \\
			\midrule
			RD4AD \cite{RD4AD} & 94.40 & 92.06 & 88.88 & 99.29&97.61&96.40 & 76.76 & 70.47 & 67.43 \\
			EDC \cite{EDC} & 95.70 & 93.22 & 90.61 & 99.59 & 98.67 & 98.00 & 90.53 & 81.94 & 86.53 \\
			INP-Former \cite{INP-Former} & 96.00 & 93.39 & 90.65 & 97.82 & 96.21 & 94.30 & \underline{90.75} & 80.50 & 83.94 \\
			EA2D \cite{EA2D} & 96.83 & 94.45 & 92.25 & \underline{99.82} & \underline{98.86} & \underline{98.28} & 88.48 & 79.29 & 84.04  \\
            Dinomaly \cite{Dinomaly} & 95.87 & 93.05 & 90.57 & 98.62 & 96.99 & 95.50 & 89.31 & 80.48 & 83.42 \\
			Recontrast \cite{ReContrast} & \underline{97.61} & \underline{95.22} & \underline{93.28} & 99.57 & 98.39 & 97.60 & 90.70 & \underline{82.16} & \underline{86.70}  \\
			\textbf{Ours} & \textbf{97.92} & \textbf{95.69} & \textbf{93.91} & \textbf{99.83} & \textbf{99.13} & \textbf{98.70} & \textbf{91.73} & \textbf{82.61}& \textbf{87.05} \\
			\bottomrule
		\end{tabular}
	%}
\end{table}

\begin{table}[!h]
	\centering
	\caption{\textcolor{r}{Performance comparison on the in-house BrainMIR dataset.}}
	\label{tab:Performance_comparison_clinical}
	\footnotesize
	\setlength{\tabcolsep}{1.5mm}
	%\resizebox{\columnwidth}{!}{%等比例缩放
            \begin{tabular}{l | ccc|cccc}
            \toprule
            \multirow{2}{*}{Method} & \multicolumn{3}{|c|}{Image-level} & \multicolumn{4}{c}{Pixel-level}\\
            \cmidrule(lr){2-4} \cmidrule(lr){5-8}
            & AUC & F1 & ACC & AUC &AP&F1&AUPRO \\
            \midrule
            RD4AD \cite{RD4AD} &73.33&75.93&67.03&89.77&5.34&11.08&62.18\\
            EDC \cite{EDC} & 88.10&83.25&80.03&96.63&27.12&33.95&84.75  \\
            EA2D \cite{EA2D} & 78.68&77.54&71.14&89.30&6.23&12.36&62.44\\
            Recontrast \cite{ReContrast} &79.06  &76.97&70.73&96.93&25.88&33.72&84.69\\
            Dinomaly \cite{Dinomaly} & \underline{91.81}&\underline{86.22}&\underline{84.31}&97.46&38.66&44.32&84.28\\
            INP-Former \cite{INP-Former} & 91.17&85.41&83.64&\underline{98.25}&\underline{43.12}&\underline{49.05}&\underline{87.92}\\
            \textbf{Ours} & \textbf{92.52}&\textbf{87.10}&\textbf{85.39}&\textbf{98.53}&\textbf{46.03}&\textbf{51.10}&\textbf{89.27} \\
            \bottomrule
            \end{tabular}
	%}
\end{table}

% \textcolor{r}{To further evaluate the efficacy of DNP-ConFormer, Fig.~\ref{fig:visualization} illustrates the anomaly maps generated across three public datasets, while Fig.~\ref{fig:visualization_clinical} provides qualitative results on our internal BrainMIR dataset. These datasets encompass a wide array of imaging modalities; the inherent structural complexity and cross-modal variability present significant hurdles for robust anomaly detection. Nevertheless, DNP-ConFormer consistently identifies pathological regions with high precision, demonstrating its adaptability to diverse clinical contexts.}

\subsection{\textcolor{r}{Ablation Study}}

To evaluate the contribution of each component, we conduct component ablation experiments focusing on encoder adaptation strategies (M0, M1, M2 and M$2^+$), sensitivity to the momentum coefficient, and Diversity-Aware Alignment loss. As summarized in Table~\ref{tab:ablation_studies}, our complete model consistently outperforms all variants across public benchmarks and the clinical BrainMIR dataset. 

Starting with \textbf{Encoder Adaptation}, a frozen encoder (M0) provides a robust baseline, whereas direct fine-tuning (M1) leads to a catastrophic performance drop ($\Delta = -37.97\%$ AUC on OCT2017), indicating severe overfitting on limited anomalous samples. While the addition of contrastive loss (M2) partially mitigates this, our momentum-based strategy (\textbf{Ours}) effectively balances domain adaptation and stability, improving AUC by $1.77\%$ and $3.27\%$ on OCT2017 and ISIC2018, respectively. 
Regarding the \textbf{Momentum Coefficient $\beta$}, we find that $\beta=0.999$ yields the optimal trade-off, achieving the highest best-epoch AUC ($90.91\%$). Lower values (e.g., $0.99$) induce training instability, shown by the larger gap between last and best AUC, while a static target ($\beta=1$) significantly limits adaptation capacity. 
Furthermore, removing the \textbf{Diversity-Aware Alignment Loss ($\mathcal{L}_{\text{daa}}$)} results in a clear performance degradation ($-0.73\%$ on APTOS2019 and $-1.06\%$ on BrainMIR), confirming its critical role in preventing prototype collapse and ensuring the learning of discriminative features across clinical scenarios.

The ablation results collectively validate that the performance gains of DNP-ConFormer stem from the synergy between its core components. While the momentum-based encoder adaptation provides a stable foundation for domain-specific feature extraction, $\mathcal{L}_{\text{daa}}$ ensures that these features remain diverse and discriminative. Together, these mechanisms ensure precise and reliable anomaly detection across a wide range of medical imaging modalities.

\begin{table}[!h]
\centering
\caption{Ablation study results}
\label{tab:ablation_studies}
% \scriptsize
\setlength{\tabcolsep}{1pt}
\newcommand{\vect}[1]{\boldsymbol{#1}}
% Left: big subtable (encoder adaptation)  
%\vspace{1mm}
\begin{minipage}[t]{0.5\textwidth}
\centering
\textbf{(a)} Encoder adaptation
%\vspace{1mm}

\begin{tabular}{@{}l rr rr@{}}
\toprule
\multirow{2}{*}{Variant} & \multicolumn{2}{c}{OCT2017} & \multicolumn{2}{c}{ISIC2018} \\
\cmidrule(lr){2-3} \cmidrule(lr){4-5}
& AUC & $\Delta_{\text{M0}}$ & AUC & $\Delta_{\text{M0}}$ \\
\midrule
M0 (Baseline) & 98.06 & -- & 88.46 & -- \\
M1 (+Unfrz.) & 60.09 & $-37.97\downarrow$ & 83.80 & $-4.66\downarrow$ \\
M2 (+CL) & 97.41 & $-0.65\downarrow$ & 85.21 & $-3.25\downarrow$ \\
Ours (+Mom.) & 99.83 & $+1.77\uparrow$ & 91.73 & $+3.27\uparrow$ \\
\bottomrule
\end{tabular}
\end{minipage}%
%\vspace{5pt}
% Right: two subtables stacked vertically (momentum above, DAA below)
\begin{minipage}[t]{0.48\textwidth}
%		\subcaptionbox{Ablation study of momentum\label{tab:ablation_m}}{%
\centering
\textbf{(b)} Momentum $\beta$
%\vspace{1mm}

\begin{tabular}{@{}l cccc@{}}
\toprule
momentum $\beta$ & 0.99 & 0.999 & 0.9999 & 1 \\
\midrule
last AUC & 85.44 & 90.00 & 88.96 & 84.58 \\
best AUC & 88.00 & 90.91 & 89.02 & 84.71 \\
\bottomrule
\end{tabular}
%\vspace{5pt} % gap between right subtables (可调)

%		\subcaptionbox{Ablation study of DAA loss on APTOS2019\label{tab:ablation_daa}}{%
\centering
\textbf{(c)} Effect of $\mathcal{L}_{\text{daa}}$
%\vspace{1mm}

\begin{tabular}{@{}l cc cc@{}}
\toprule
\multirow{2}{*}{Strategy} & \multicolumn{2}{c}{APTOS2019} & \multicolumn{2}{c}{BrainMIR}\\
\cmidrule(lr){2-3}  \cmidrule(lr){4-5} 
& AUC & $\Delta$ &  AUC & $\Delta$ \\
\midrule
w \;\;\:$\mathcal{L}_\text{daa}$ &95.97  &  &92.52& \\
w/o $\mathcal{L}_\text{daa}$&95.24 &${-0.73}\downarrow$   &91.46& ${-1.06}\downarrow$\\
\bottomrule
\end{tabular}
\end{minipage}
\end{table}

%\section{Conclusions}
%\label{Conclusions}
%The strong performance of DNP-ConFormer across diverse medical imaging datasets highlights its potential as a unified framework for a wide range of anomaly detection tasks. Its robustness on both structured modalities (e.g., OCT, CT) and unstructured ones (e.g., fundus and dermoscopic images) demonstrates that the DNPs effectively preserve rich local semantics and prevent prototype collapse, while the dual-branch comparative reconstruction ensures powerful global representation.
%
%\textbf{Limitations and future work.} 
%Nevertheless, certain limitations remain. On the visually complex ISIC2018 dataset, performance degradation is observed across all models, including DNP-ConFormer, indicating the need for more refined modeling of fine-grained and highly variable skin lesion features. Moreover, the current framework assumes clean training conditions. Future work could explore the robustness of DNP-ConFormer under label noise.
%
%\textbf{Summary.} 
%DNP-ConFormer presents a promising and generalizable approach to medical anomaly detection by integrating diversity-aware prototype modeling with a contrastive architecture, significantly improving detection performance.

\section{Conclusion}
The robust performance of DNP-ConFormer across diverse datasets establishes its potential as a unified framework for medical anomaly analysis. Its stability on both structured (e.g., OCT, MIR) and unstructured (e.g., fundus, dermoscopic) modalities confirms that the DNPs effectively preserve rich local semantics and preclude prototype collapse, while the dual-branch comparative reconstruction facilitates powerful global representation. Furthermore, the superior results on the in-house BrainMIR dataset underscore its reliability for deployment in real-world clinical settings.
Overall, DNP-ConFormer provides a robust and generalizable framework, offering a high-performance alternative for diverse medical anomaly analysis tasks.

\bibliographystyle{splncs04}
\bibliography{ref}

\end{document}